\documentclass{article}

\PassOptionsToPackage{numbers, compress}{natbib}

 \usepackage[preprint]{nips_2018}

\usepackage[T1]{fontenc}    
\usepackage{url}            
\usepackage{booktabs}       
\usepackage{amsfonts}       
\usepackage{nicefrac}       
\usepackage{microtype}      
\usepackage{tabularx}
\usepackage{algorithm}
\usepackage{algorithmic}
\usepackage{epsfig}
\usepackage{amsmath}
\usepackage{multirow}
\usepackage{subfigure}

\newtheorem{theorem}{\textbf{Theorem}}

\DeclareMathOperator*{\VEC}{vec}

\DeclareMathOperator*{\T}{T}

\def\Xb{{\textbf{X}}}

\def\Wb{{\textbf{W}}}
\def\Mb{{\textbf{M}}}

\title{Towards Discrete Solution: A Sparse Preserving Method for Correspondence Problem}

\author{
  Bo Jiang \\
  School of Computer Science and Technology \\
  Anhui University\\
  Hefei, China \\
  \texttt{jiangbo@ahu.edu.cn} \\
}

\begin{document}

\maketitle

\begin{abstract}
\noindent Many problems of interest in computer vision can be formulated as
a problem of finding consistent correspondences between two feature sets.
Feature correspondence (matching) problem with one-to-one mapping constraint is usually formulated as an Integral Quadratic Programming (IQP) problem
with permutation (or orthogonal) constraint.
Since it is NP-hard, relaxation models are required.
One main  challenge for optimizing IQP matching problem is how to incorporate the discrete one-to-one mapping (permutation) constraint
in its quadratic objective optimization.
In this paper, we present a new relaxation model, called Sparse Constraint Preserving Matching (SPM), for IQP matching problem.
SPM is motivated by our observation that the discrete permutation constraint can be well encoded via a sparse constraint.
Comparing with traditional relaxation models, SPM can incorporate the discrete one-to-one mapping constraint straightly
via a sparse constraint and thus provides a tighter relaxation for original IQP matching problem.
A simple yet effective update algorithm has been derived to solve the proposed SPM model.
Experimental results on several feature matching tasks demonstrate the effectiveness and efficiency of SPM method.

\end{abstract}

\section{Introduction}

Many problems of interest in computer vision and pattern recognition field can be formulated
as a problem of finding consistent correspondences (matching) between two sets of features.
Feature matching problem with one-to-one mapping constraint can usually be formulated as an
Integral Quadratic Programming (IQP) problem with permutation (or orthogonal) constraint. 
There are two main challenges for optimizing this IQP matching problem: (1) The
quadratic objective function of this problem is generally non-convex and prone to local optima;
(2) The discrete one-to-one mapping constraint can not be naturally/easlily incorporated into the quadratic objective optimization.

 In the past decade, various relaxation methods have been developed to find an approximate solution
 for this problem.
One kind of relaxation methods is to use a two-stage optimization procedure
 which first optimizes the problem in a continuous domain (e.g. bistochastic domain) by
  relaxing the integer constraint and then utilizes a post-optimization/discretization stage to compute the  final discrete solution~\cite{RWGM,GA,SMAC,SM}.
One main limitation is that the required post-optimization stage is generally independent of the  matching objective optimization
and thus usually leads to weak local optima.
Another kind of methods is trying to optimize the IQP matching problem directly in the discrete domain and thus can
obtain the discrete solution directly for the problem.
For example,
Leordeanu et al.~\cite{IPFP} proposed an integer projected matching method (IPFP) to obtain a discrete binary solution for the problem.
Zhou et al.~\cite{FGM} proposed a factorized graph matching (FGM) which optimizes the IQP
problem using a convex-concave relaxation technique and returns a discrete solution for the problem.
Kamil et al.~\cite{Tabu} proposed to use a discrete tabu search technique for the problem.
Both FGM~\cite{FGM} and Tabu~\cite{Tabu} suffer from high computational complexity.
Recently, Jiang et al.~\cite{NOGM} proposed to solve the  matching problem with nonnegative
orthogonal constraint approximately and thus can return an approximate discrete solution.
In addition to the above continuous and discrete methods, sparse relaxation models have also been
explored for feature correspondence problem~\cite{GameM3,GameM2,LSM,EnetM}.
The main benefit of these sparse methods is that they can obtain an optimal discrete solution for the problem
by using a very simple algorithm.
However, one important drawback is that they generally fail to incorporate the one-to-one mapping constraint in its optimization process.

In this paper,  we propose a new relaxation model, called Sparse Constraint Preserving Matching (SPM), for the IQP matching problem with
one-to-one mapping constraint.
SPM is motivated by our new observation that the discrete permutation constraint in IQP matching problem can be well represented (or
encoded) by a structure sparse constraint. Comparing with original permutation constraint, the sparse constraint is easier to
 impose by borrowing the sparse relaxation techniques from machine learning field.
To the best of our knowledge, this particular observation has not been explored or emphasized before.
One main benefit of SPM is that it incorporates the discrete one-to-one mapping constraint naturally
via a simple structure sparse constraint in its optimization procedure.
Also, a simple yet very effective update algorithm has been derived to solve SPM model.
The convergence and optimality properties of the algorithm are guaranteed in theoretical.
Experimental results on several feature correspondence datasets demonstrate the effectiveness and efficiency of SPM method.

\section{Problem Formulation and Related Works}

\subsection{Problem formulation}

Given two feature sets $\mathcal{M}$, containing $n_M$ model features and $\mathcal{D}$, containing $n_D$ data features,
the aim of feature matching is to find consistent correspondences between two sets of features.
Let $\Wb_{ij,kl}$ measure how compatible the feature pair $(i, k)\in \mathcal{M}$ is with the feature pair $(j, l)\in \mathcal{D}$  in both local appearance and pair-wise geometry constraint (or any other type of pairwise relationship).
Let $\Xb\in \{0,1\}^{n\times n}$ ($n=n_M=n_D$)\footnote{Here, we focus on equal size feature set matching. For the sets
with different sizes, one can transform it to equal size problem by
adding dummy features to the smaller set.} denote the correspondence indication matrix, in which $\Xb_{ij}=1$ implies that
feature $i\in\mathcal{M}$ corresponds to feature $j\in\mathcal{D}$, and $\Xb_{ij}=0$ otherwise. Then, the one-to-one feature correspondence problem is
generally formulated an Integeral Quadratic Programming (IQP) problem with permutation constraint, i.e.,
\begin{align}\label{EQ:GMIQP}
&  \max_{\Xb}\ \VEC{(\Xb)}^{\T} \Wb \VEC{(\Xb)} \nonumber \\
& s.t.\ \ \  \Xb\textbf{1}  = \textbf{1},\Xb^{\T}\textbf{1} = \textbf{1}, \Xb_{ij} \in \{0,1\}.
\end{align}
%
where $\textbf{1}=(1,1,\cdots 1)^{\T}\in \mathbb{R}^{n\times 1}$ and $\VEC{(\Xb)}=(\Xb_{11},\Xb_{21},\cdots \Xb_{nn})^{\T} \in \mathbb{R}^{n^2\times 1}$.
The permutation constraint ensures the one-to-one mapping between two feature sets,
i.e., there exists exactly one entry 1 in each column and each row of $\textbf{X}$.
In this paper, we use  the terms matching  and correspondence interchangeably.

\subsection{Related works}

It is known that, the above IQP problem (Eq.(1)) is NP-hard. Thus, relaxation models and algorithms are developed to find approximate solutions.
One kind of  relaxation models is to explore sparse constraint on related solution and thus can maintain the discrete constraint approximately.
%

\textbf{Game Theoretic Matching (GTM)}:
From game theoretic perspective, Albarelli et al. \cite{GameM} proposed a simple
sparse relaxation model by adding a $\ell_1$-norm constraint on related solution, i.e.,
\begin{equation}\label{EQ:GameM}
 \max_{\Xb}\  \ \VEC{(\Xb)}^{\T} \textbf{W} \VEC{(\Xb)} \  \ \ \ s.t.\  \|\Xb\|_1 =1, \Xb\geq 0.
\end{equation}
where $\|\Xb\|_1 = \sum_{ij}|\Xb_{ij}|$. It can generate a sparse solution for the problem due to $\ell_1$-norm constraint. 
%

\textbf{Elastic Net Matching (EnetM)}:
Rodol\`a et al. \cite{EnetM} introduced a more stable relaxation matching model by imposing an elastic net constraint on related solution as,
\begin{align}\label{EQ:EnetM}
 \max_{\Xb}\  \  \ \VEC{(\Xb)}^{\T} \Wb \VEC{(\Xb)} \ \   \ s.t.\ \ \  (1-\epsilon)\|\Xb\|_1 + \epsilon\|\Xb\|^2_2 =1, \Xb\geq 0,
\end{align}
where $\epsilon \in [0,1]$ is a balanced parameter.

\textbf{Local Sparse Matching (LSM)}:
Jiang et al. \cite{LSM} recently introduced a more accurate relaxation  model for correspondence problem by developing a local sparse matching model as,
\begin{align}\label{EQ:EnetM}
 \max_{\Xb}\  \  \ \VEC{(\Xb)}^{\T} \Wb \VEC{(\Xb)} \ \  \ \ s.t.\ \ \  \|\Xb\|_{1,2} =1, \Xb\geq 0
\end{align}
where $\|\Xb\|_{1,2} = \big(\sum_i(\sum_j |\Xb_{ij}|)^2\big)^{1/2}$ is a $\ell_{1,2}$-norm function.
Clearly, LSM~\cite{LSM} incorporates the matching constraint more strongly than GTM \cite{GameM}  and EnetM~\cite{EnetM}.

One important drawback for the above relaxation models is that they generally fail to consider the one-to-one mapping constraint in its relaxation
and optimization process.

\section{Sparse Constraint Preserving Matching}

In this section, we first propose our relaxation model for the original IQP matching problem (Eq.(\ref{EQ:GMIQP})),
called Sparse Constraint Preserving Matching (SPM). 
Then, we develop a simple and effective update algorithm to solve the proposed SPM model.


\subsection{SPM Model}
Our SPM model is motivated by an observation that the discrete permutation constraint in IQP matching problem (Eq.(\ref{EQ:GMIQP}))
can be approximately represented by a \textbf{structure sparse constraint}.
Particularly, from sparse (number of non-zero entries) aspect,
the above permutation/orthogonal constraint in Eq.(\ref{EQ:GMIQP}) can be equivalently represented as that
\emph{there exists only one non-zero entry (equals to 1) in each row} $\Xb_{i\cdot}$ \emph{and each column} $\Xb_{\cdot j}$ \emph{of solution matrix}.

Obviously, this exact sparse constraint is also combinatorial and thus NP hard. However, comparing with permutation constraint,
 sparse constraint is much easier to implement.
Our aim is to develop an alternative continuous way to encode this sparse constraint approximately by adopting a dual $\ell_{1,2}$-norm constraint.
%
 %
 Formally, our SPM model is formulated as,
\begin{align}\label{EQ:SPM}
&  \max_{\Xb}\ \VEC{(\Xb)}^{\T} \Wb \VEC{(\Xb)} \nonumber \\
& s.t.\ \ \sum\nolimits^n_{i=1} \|\Xb_{i\cdot}\|^2_1 + \sum\nolimits^n_{j=1} \|\Xb_{\cdot j}\|^2_1=1, \Xb_{ij}\geq 0.
\end{align}
where $\|\cdot\|_1$ denotes the $\ell_1$-norm function. The notation $\Xb_{i\cdot}$ and $\Xb_{\cdot j}$ denote the $i$-th row and $j$-th column of solution matrix $\Xb$, respectively.

Note that, (1) the $\ell_1$-norm constraints on each row and column of matrix $\Xb$ encourage the sparse constraint on each row and column, i.e., there exist a few number of non-zero entries in each row and column.
(2) The $\ell_2$-norm constraints on rows and columns encourage that there is no zero row and column in $\Xb$ which encourage that the number of non-zero entries in each row and column is at least one.
Based on (1) and (2), one can see that the optimal solution of SPM model is
row-column sparse and thus approximate orthogonal, as shown in \S 3.3.
%
\subsection{Optimization}

We propose a simple yet effective update algorithm to solve the proposed SPM problem.
First, since $\Xb_{ij}\geq 0$, using matrix representation  we can rewrite Eq.(\ref{EQ:SPM}) more compactly as
\begin{align}\label{EQ:SR1}
&  \max_{\Xb}\ \VEC{(\Xb)}^{\T} \Wb \VEC{(\Xb)} \nonumber \\
& s.t.\ \  \textbf{1}^{\T}(\Xb^{\T}\Xb +\Xb\Xb^{\T}) \textbf{1} = 1, \Xb_{ij}\geq 0.
\end{align}
where $\textbf{1} = (1,1\cdots 1)^{\T}\in \mathbb{R}^{n\times 1}$.

Then, we derive an algorithm to find a local optimal solution for SPM problem.
Formally, given an initial solution $\Xb^{(0)}$, we propose to iteratively update the current solution $\Xb^{(t)}$ as follows,
%
\begin{equation}\label{EQ:Update}
\textbf{X}^{(t+1)}_{ij} = \Xb^{(t)}_{ij}\Big[\frac{\Mb^{(t)}_{ij}}{\alpha^{(t)} (\textbf{1}\textbf{1}^{\T}\Xb^{(t)} + \Xb^{(t)} \textbf{1}\textbf{1}^{\T} )}_{ij}\Big]^{1/2},
\end{equation}
%
where matrix $\Mb^{(t)}\in\mathbb{R}^{n\times n}$ is the matrix form of the vector $[\Wb\VEC{(\Xb^{(t)})}]$, i.e., $\VEC{(\Mb^{(t)})} = \Wb \VEC{(\Xb^{(t)})}$, and $\alpha^{(t)}$ is computed as,
\begin{equation}
\alpha^{(t)} = \VEC{(\Xb^{(t)})}^{\T} \Wb \VEC{(\Xb^{(t)})}.
\end{equation}
The iteration starts with a nonnegative initial solution $\textbf{X}^{(0)}$ and is repeated until convergence. Since $\Wb$ is a real nonnegative matrix for the feature correspondence problem, the nonnegativity of $\Xb^{(t)}$ is always guaranteed in each iteration.

\textbf{Complexity analysis.} The main computational complexity in each iteration is on computing $\Wb \VEC{(\Xb^{(t)})}$.
Assume $\Wb \in \mathbb{R}^{N\times N}$, the main complexity for SPM is less than $O(\mathrm{M}N^2)$,
where $\mathrm{M}$ is the maximum iteration.
Empirically, the algorithm converges quickly and the average maximum iteration $\mathrm{M}$ is generally less than 200.

%
%
%
%

\subsection{Illustration}

Figure 1(a) shows solution $\Xb^{(t)}$ across different iterations.
One can note that as the iteration increases,
the solution matrix $\Xb^{(t)}$ becomes more and more sparse and approximates to a
permutation matrix closely at convergence, which indicates the ability of SPM model on
maintaining the discrete one-to-one mapping constraint in its optimization.
Figure 1(b) shows the objective, sparsity and orthogonality of solution $\Xb^{(t)}$.
We can observe that: (1) The objective of $\Xb^{(t)}$ increases and
converges after some iterations, demonstrating the convergence of the proposed update algorithm.
(2) As the iteration increases, the number of non-zero entries in $\Xb^{(t)}$ decreases, which shows the
sparse property of SPM model.
(3) As the iteration increases, the orthogonality (mean value of off-diagonal entries in $\Xb\Xb^{\T}$) of $\Xb^{(t)}$
decreases and approximates to zero at convergence, which is consistent with the demonstration of Figure 1(a).
%
\begin{figure*}[htp]
  \centering
\includegraphics[width=0.9\textwidth]{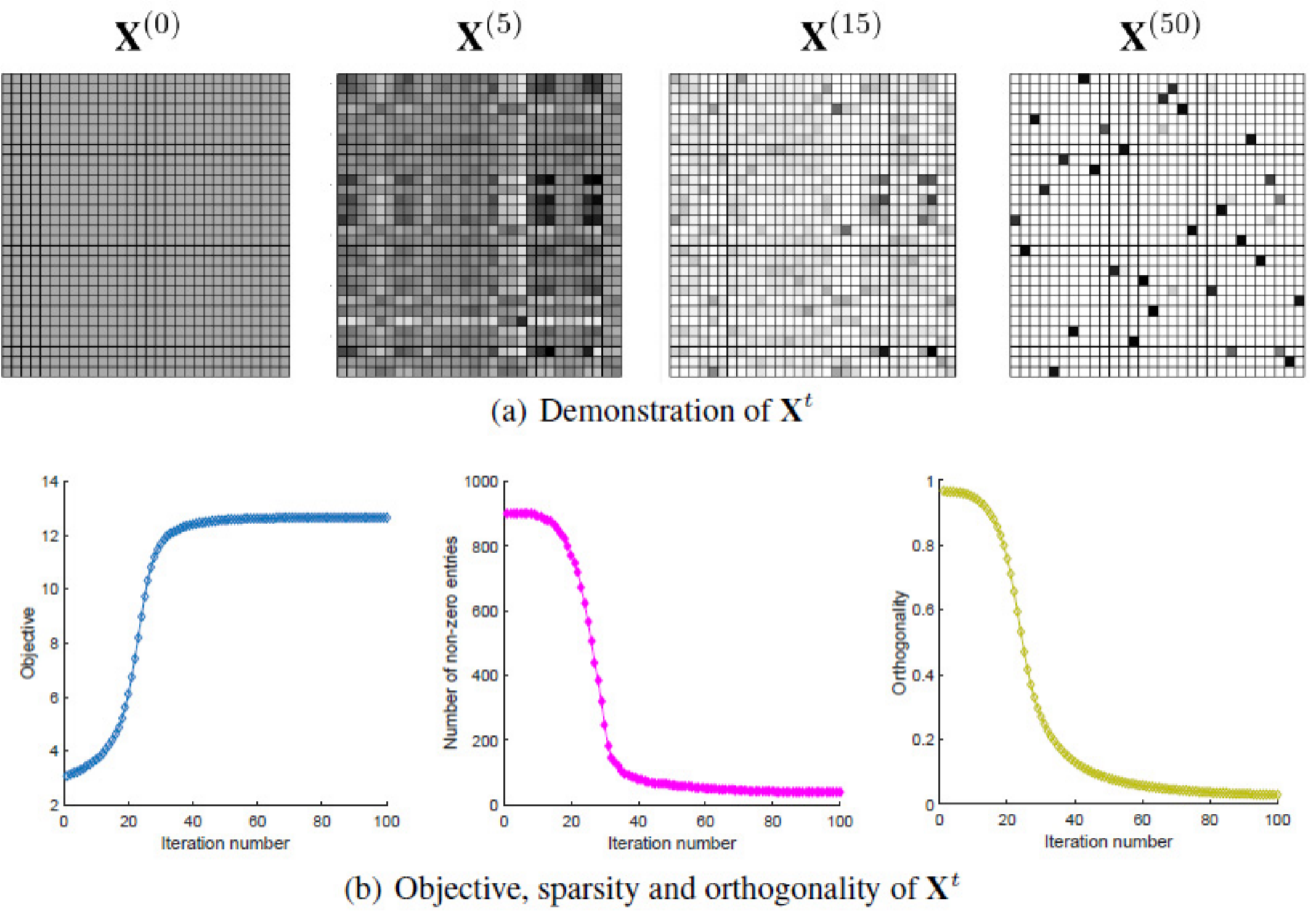}
\caption{Illustration of SPM method and solution.}
\end{figure*}

\section{Theoretical Analysis}
In this section, we show the convergence and optimality of the update algorithm.
%
\begin{theorem}
Under the update algorithm of Eq.(\ref{EQ:Update}), the following Lagrangian function $\mathcal{L}(\textbf{X})$  is monotonically increasing.
\end{theorem}
\begin{equation}\label{EQ:Lagrangian}
\mathcal{L}(\Xb) = \VEC{(\Xb)}^{\T}\Wb\VEC{(\Xb)} - \alpha \big(\textbf{1}^{\T}(\Xb^{\T}\Xb+\Xb\Xb^{\T})\textbf{1} - 1\big)
\end{equation}
\begin{theorem}
The converged solution obtained by update rule (Eq.(\ref{EQ:Update})) satisfies the first-order KKT optimality condition.
\end{theorem}

\section{Experiments}

To evaluate the effectiveness of the proposed SPM method, we test the proposed method on some feature correspondence tasks.
We compare our SPM with some other methods including (i)
continuous domain method GA \cite{GA}, SM \cite{SM}, SMAC \cite{SMAC} and RRWM \cite{RWGM};
(ii) discrete optimization method IPFP \cite{IPFP}, FGM \cite{FGM}, NOGM \cite{NOGM} and Tabu \cite{Adamczewski2016Discrete};
(iii) sparse relaxation method LSM \cite{LSM}.
We only compare our SPM with LSM \cite{LSM} because it obtains obviously better performance than other sparse method GameM \cite{GameM} and EnetM \cite{EnetM}.
We implement our method on a PC with an Intel i7 4.0GHz CPU and 16GB RAM, and the proposed algorithm is implemented in MATLAB R2014b.

\begin{figure*}[htp]
  \centering
\includegraphics[width=0.97\textwidth]{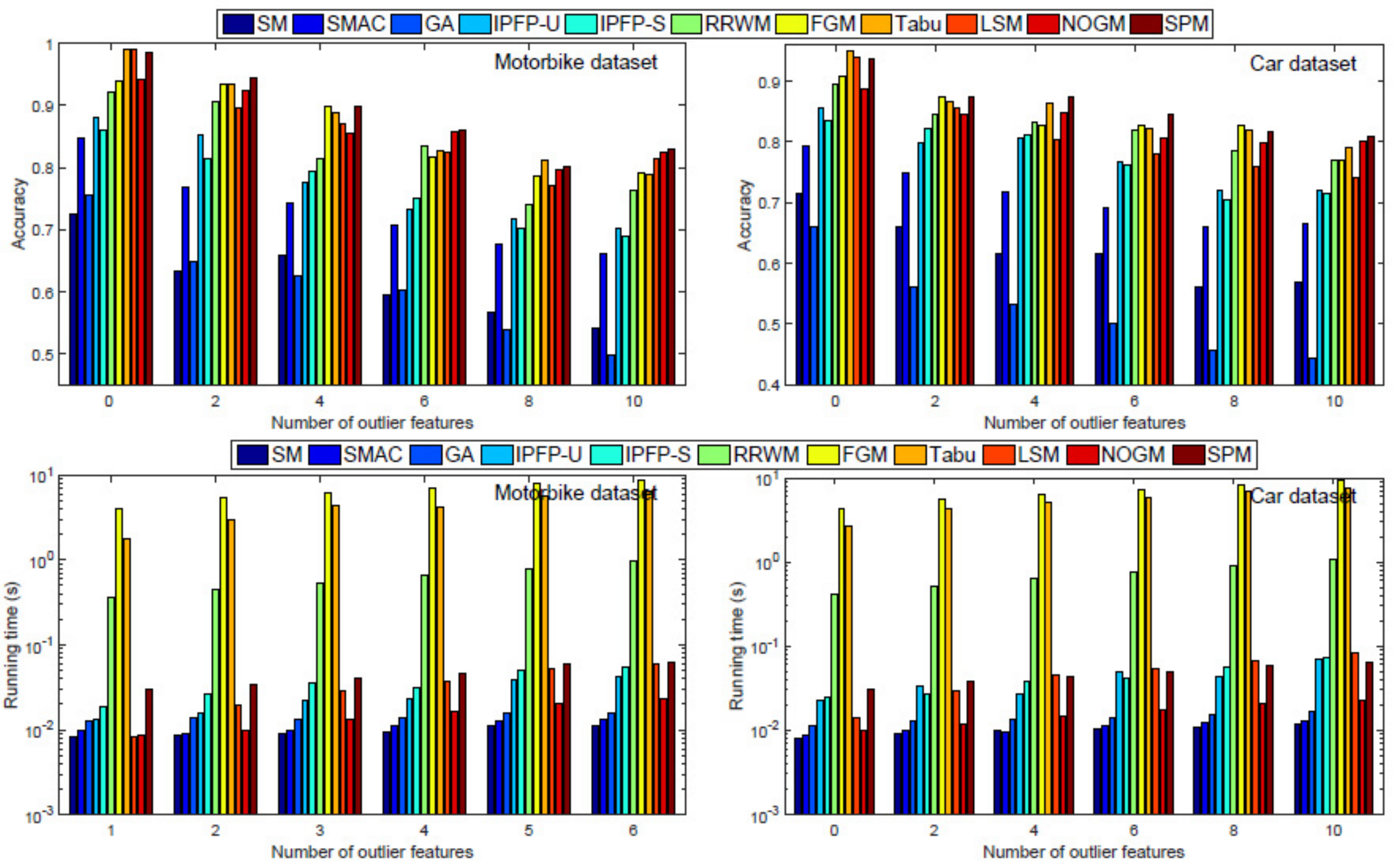}
\caption{ Comparison of matching accuracy results of different feature matching methods on Pascal 2007 dataset (LEFT: Motorbike dataset; RIGHT: Car dataset).}
\end{figure*}
%

%
%
\subsection{Image matching  on  Pascal 2007 dataset}
We first evaluate our SPM method on the image dataset \cite{Leordeanu2012Unsupervised} selected from Pascal 2007 \footnote{http://www.pascalnetwork.org/challenges/VOC/voc2007/workshop/index.html}.
This dataset contains 50 image pairs including 30 pairs of car images and 20 pairs of
motorbike images.
For each image pair, the feature points and ground-truth correspondences
were manually marked.
Following the experimental setting \cite{FGM}, we use the Delaunay triangulation to build a graph for each image whose nodes
denote the feature point and edges represent the distance $d_{ik}$ and absolute angle $\theta_{ik}$ of edges. 
%
%

\textbf{Effectiveness:} 
To test the performance against outlier features, we have randomly added 0-10 outlier features for each image pair.
The overall results of matching accuracy across different outlier features are summarized in Figure 2.
Here, we can note that, SPM generally performs better than other related methods, which demonstrates the
effectiveness of the proposed SPM method. More detailly,
(1) SPM obtains obviously better performance than traditional continuous relaxation method GA~\cite{GA}, SMAC~\cite{SMAC} and RRWM~\cite{RWGM}.
This clearly demonstrates the effectiveness of SPM relaxation model by 
incorporating the discrete one-to-one mapping constraint in matching problem optimization.
(2) SPM performs better than some discrete optimization methods, such as IPFP~\cite{IPFP}, FGM~\cite{FGM}, Tabu~\cite{Tabu} and NOGM~\cite{NOGM}.
Comparing with these methods, SPM solution is guaranteed to be KKT optimal and thus performs more optimally and effectively.
(3) Comparing with LSM~\cite{LSM}, SPM provides a tighter relaxation for original IQP matching problem and thus generally outperforms LSM~\cite{LSM},
especially on solving the matching problem with outlier features.


\textbf{Efficiency:} 
Figure 3 shows the comparison of running time across different number of outlier features.
 One can note that SPM method performs obviously faster than RRWM~\cite{RWGM}, FGM~\cite{FGM} and Tabu \cite{Adamczewski2016Discrete} methods which are
 the most competitive methods with our SPM.
Particularly, the speed of SPM is about 100 times faster than FGM~\cite{FGM}, at least 50 times faster than Tabu \cite{Adamczewski2016Discrete} and
about 10 times faster than RRWM~\cite{RWGM} on the setting of this dataset.
This further demonstrates the efficiency of the proposed SPM method.
\begin{figure*}[htp]
  \centering
\includegraphics[width=0.99\textwidth]{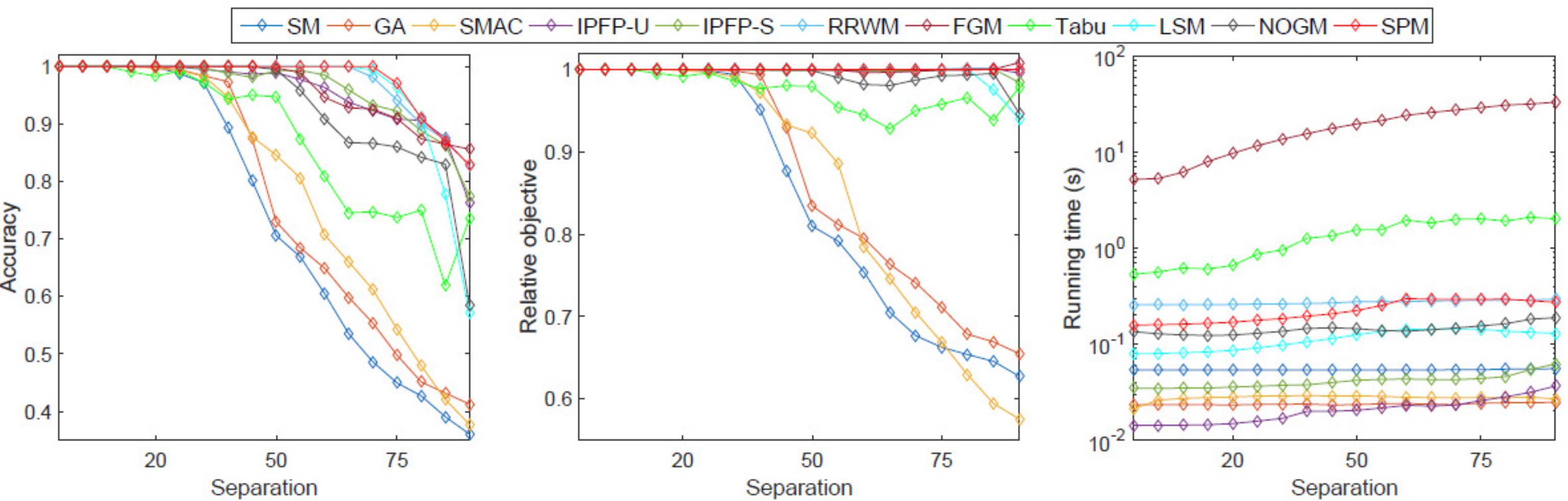}
\caption{ Comparison of matching accuracy results of different feature matching methods on Pascal 2007 dataset (LEFT: motorbike dataset; RIGHT: Car dataset).}
\end{figure*}
\subsection{Image sequence dataset}

In this section, we perform feature matching on CMU "hotel" sequence dataset which has been widely used for feature matching algorithm evaluation~\cite{RWGM,LearnGM,SpetralEmbedding},.
For CMU "hotel" sequence, there are 101 images of a toy house captured from moving viewpoints.
For each image, 30 landmark points were manually marked with known correspondences.
We have matched all images separated by 5, 10 $\cdots$ 75 and 80 frames and computed the average performances per separation gap.
%
Figure 4 summarizes the performance results.
It is noted that SPM outperforms the other comparing methods in  matching accuracy and general returns high
relative objective score, which demonstrates the effectiveness of SPM method.
Also, SPM performs faster than some other competitive methods, which further demonstrates
the efficiency of SPM method.

\section{Conclusions and Future work}
This paper makes two main contributions.
First, we show that the discrete one-to-one mapping constraint in IQP correspondence problem can be well encoded by a structure sparse constraint.
Second, based on this observation, we propose a novel tighter relaxation model SPM  for IQP matching problem with one-to-one mapping constraint.
The main benefit of SPM is that it can incorporate the discrete one-to-one mapping constraint naturally in the quadratic objective optimization procedure.
A simple and effective algorithm has been derived to find a local optimal solution for SPM model.

Note that, the proposed sparse preserving and relaxation model is not limited for solving feature matching problem. It can also been adapted for solving
 some other similar problems, such as MAP, clustering and so on.
In our future, we will also explore some more effective optimization strategy, such as path-following strategy, to obtain a more optimal solution for SPM model.
  {\small
\bibliographystyle{ieee}
\bibliography{nmfgm}
}

\end{document}